\documentclass[10pt,twocolumn,letterpaper]{article}

\usepackage[pagenumbers]{wacv} % To force page numbers, e.g. for an arXiv version

\usepackage{cuted}   % for wide environments in two-column mode
\usepackage{capt-of} % for captionof

\newenvironment{teaserfigure}{
    \begin{strip}
    \vspace*{-1.5cm} 
    \centering
}{
    \end{strip}
}

\usepackage{multirow}
\definecolor{wacvblue}{rgb}{0.21,0.49,0.74}
\usepackage[pagebackref,breaklinks,colorlinks,allcolors=wacvblue]{hyperref}

\def\wacvPaperID{xxxx} % *** Enter the WACV Paper ID here
\def\confName{xxx}
\def\confYear{xxx}

\title{\LARGE ToonifyGB: StyleGAN-based Gaussian Blendshapes for 3D Stylized Head Avatars}

\author{
Rui-Yang Ju$^{1}$ \quad Sheng-Yen Huang$^1$ \quad Yi-Ping Hung$^1$ \\
$^1$Graduate Institute of Networking and Multimedia, National Taiwan University, Taiwan \\
{\tt\small jryjry1094791442@gmail.com, d12944001@csie.ntu.edu.tw, hung@csie.ntu.edu.tw} \\
{\url{https://ruiyangju.github.io/ToonifyGB}}
} 
\begin{document}
\maketitle

\begin{teaserfigure}
\centering
\includegraphics[width=\linewidth]{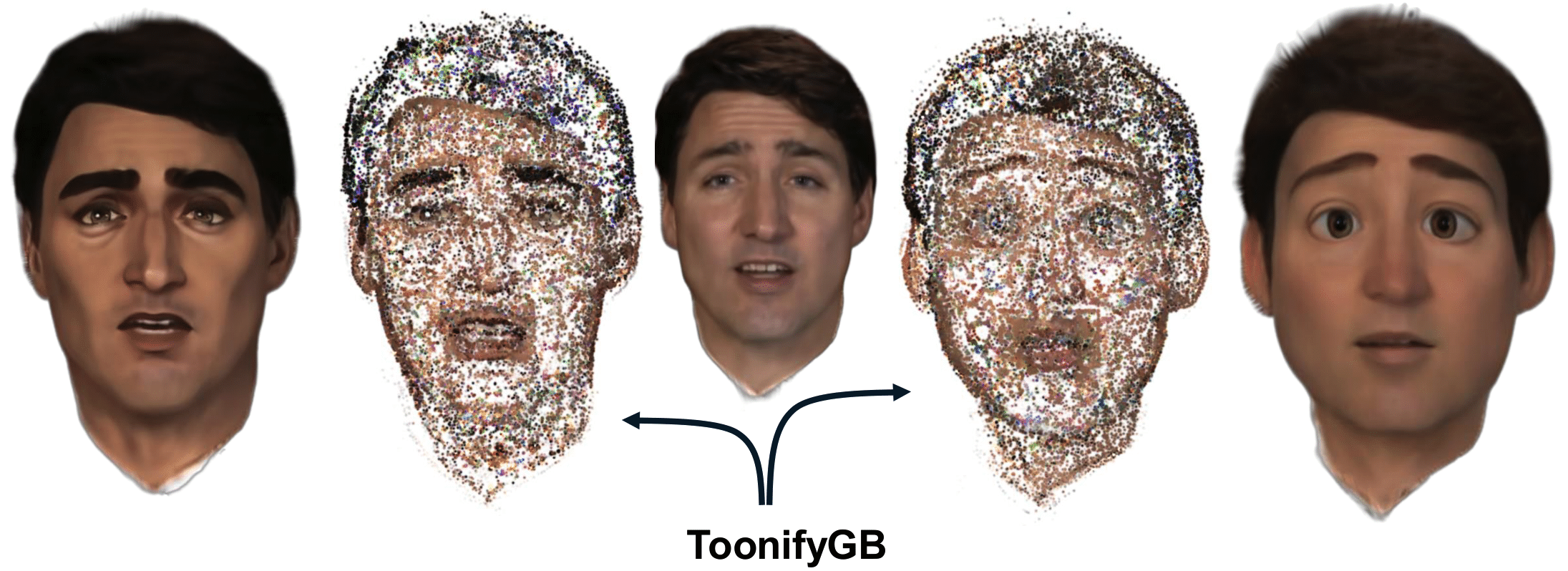}
\captionof{figure}{\textbf{ToonifyGB}: 
We propose an efficient two-stage framework that employs an improved StyleGAN to generate stylized head videos from input video frames and synthesize the corresponding 3D avatars using Gaussian blendshapes. 
Our method supports real-time synthesis of stylized avatar animations (with 50k Gaussians for the neutral model and 14k Gaussians for the mouth interior) in diverse styles such as Arcane and Pixar.}
\end{teaserfigure}

\begin{abstract}
The introduction of 3D Gaussian blendshapes has enabled the real-time reconstruction of animatable head avatars from monocular video. 
Toonify, a StyleGAN-based method, has become widely used for facial image stylization.
To extend Toonify for synthesizing diverse stylized 3D head avatars using Gaussian blendshapes, we propose an efficient two-stage framework, ToonifyGB.
In Stage~1 (stylized video generation), we adopt an improved StyleGAN to generate the stylized video from the input video frames, which overcomes the limitation of cropping aligned faces at a fixed resolution as preprocessing for normal StyleGAN.
This process provides a more stable stylized video, which enables Gaussian blendshapes to better capture the high-frequency details of the video frames, facilitating the synthesis of high-quality animations in the next stage.
In Stage~2 (Gaussian blendshapes synthesis), our method learns a stylized neutral head model and a set of expression blendshapes from the generated stylized video.
By combining the neutral head model with expression blendshapes, ToonifyGB can efficiently render stylized avatars with arbitrary expressions. 
We validate the effectiveness of ToonifyGB on benchmark datasets using two representative styles: Arcane and Pixar.
\end{abstract}

\section{Introduction}
With the advancement of 3D head reconstruction technologies, individuals can now personalize unique avatars for telepresence and virtual/augmented reality applications, which serve as a crucial foundation for the rise of the metaverse.
Considering user preferences and privacy concerns, the creation of stylized avatars has become an important research topic.
Toonify~\cite{pinkney2020resolution}, a StyleGAN-based method, was designed for 2D facial image stylization, presenting the potential of translating real portraits into stylized 2D images.
While such methods focus on 2D images, recent advances in 3D head reconstruction have mainly targeted photo-realistic avatars. 
In contrast, stylized 3D head avatars emphasize personal identity and the faithful transfer of artistic styles.

Blendshapes are an efficient facial animation representation that synthesize continuous and high-quality expressions by blending a set of 3D meshes, each corresponding to a specific facial expression.
These facial shapes are synthesized by linearly blending the basis meshes using weighting coefficients.
With the introduction of Neural Radiance Fields (NeRF)~\cite{mildenhall2021nerf}, Gao \emph{et al.}~\cite{gao2022reconstructing} and Zheng \emph{et al.}~\cite{zheng2022avatar} incorporated the blendshape concept into NeRF, enabling avatar animation through a group of NeRF blendshapes that are linearly blended.
Furthermore, the recently proposed 3D Gaussian Splatting (3DGS)~\cite{kerbl20233d} significantly improved rendering efficiency and achieved higher-quality head reconstruction, outperforming NeRF in both speed and quality.
Building on this, 3D Gaussian Blendshapes (3DGB)~\cite{ma20243d} successfully integrated blendshapes with Gaussian splatting, achieving real-time rendering and state-of-the-art performance in head reconstruction.

In contrast to previous works focused on photo-realistic 3D head avatar reconstruction, we propose ToonifyGB, a two-stage framework for synthesizing and animating 3D stylized head avatars.
Given monocular video frames, Stage~1 adopts an improved StyleGAN to generate a more stable and less jittery stylized video, without requiring fixed resolution or pre-aligned face cropping.
In Stage~2, we build upon 3DGB to learn a neutral head model and a set of expression blendshapes, each represented as 3D Gaussians.
Finally, by incorporating a facial tracker~\cite{zielonka2022towards}, ToonifyGB uses the tracked motion parameters to animate 3D stylized head avatars.

The contributions of this work are as follows:
\begin{itemize}
\item We propose ToonifyGB, an efficient two-stage framework that synthesizes 3D stylized head avatars from monocular videos using Gaussian blendshapes, supporting diverse styles with real-time animation.
\item We demonstrate that reducing per-frame jitter in the generated video enables Gaussian blendshapes to better capture high-frequency details, thereby improving the quality of 3D stylized head avatar animations.
\item To the best of our knowledge, this work is the first to synthesize 3D stylized head avatars using Gaussian blendshapes.
\end{itemize}

\begin{figure*}[t]
\centering
\includegraphics[width=\linewidth]{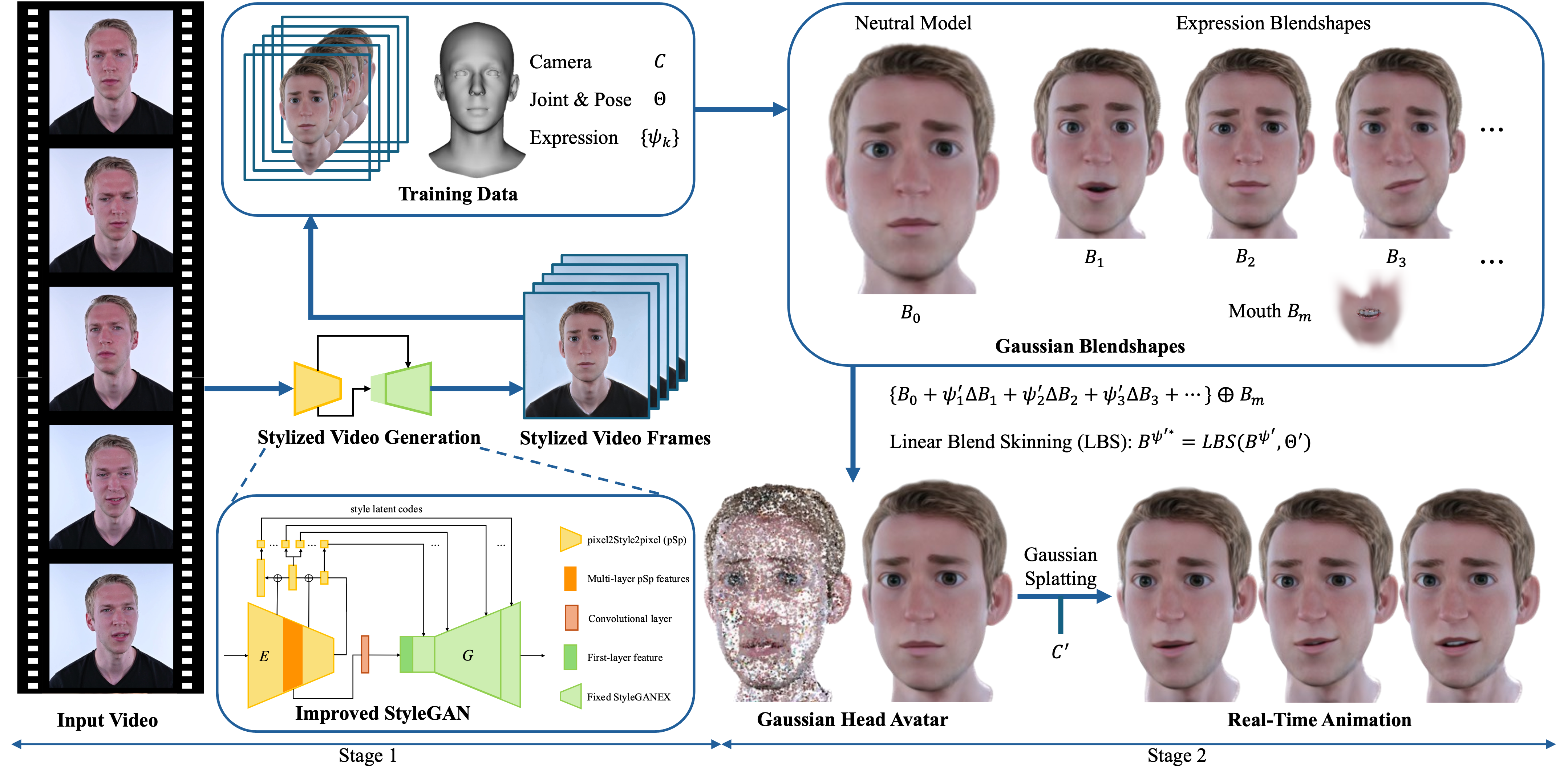}
\caption{\textbf{Pipeline:} 
Our ToonifyGB framework consists of two stages:
Stage 1 involves the generation of stylized videos, and Stage 2 focuses on the synthesis of 3D stylized head avatars using Gaussian blendshapes.}
\label{fig:pipeline}
\end{figure*}

\section{Related Work}
\subsection{StyleGAN and Toonify}
StyleGAN~\cite{karras2019style,karras2020analyzing} has been widely used to generate realistic facial images across diverse styles. 
Inversion of StyleGAN enables projecting real facial images into its latent space, allowing subsequent edits such as adding glasses or changing hairstyles or age~\cite{abdal2019image2stylegan,patashnik2021styleclip}.
To enhance inversion efficiency, methods such as pSp~\cite{richardson2021encoding} and e4e~\cite{tov2021designing} employ encoders to directly project target faces into their corresponding latent codes.
However, these methods often struggle to reconstruct fine image details, resulting in unsatisfactory reconstruction quality. 
To address these limitations, ReStyle~\cite{alaluf2021restyle} and HFGI~\cite{wang2022high} improve reconstruction fidelity by respectively predicting latent code residuals and correcting intermediate features.
Nevertheless, these methods remain limited to aligned and cropped facial images for effective editing and reconstruction.

Recently, researchers~\cite{pinkney2020resolution,ojha2021few,jang2021stylecarigan,yang2022pastiche,gal2022stylegan} have explored the use of StyleGAN for target-domain image generation through transfer learning. 
Among these works, Toonify~\cite{pinkney2020resolution} fine‑tunes the trained generator to blend realistic textures with toonified facial structures. 
In addition to image editing, StyleGAN has also been widely applied to video editing. 
Related studies have focused on enhancing video editing performance by employing temporal correlations in low‑dimensional latent codes~\cite{fox2021stylevideogan}, disentangling identity from facial attributes~\cite{yao2021latent}, incorporating sketch‑based branches~\cite{liu2022deepfacevideoediting}, and tuning the generator to maintain temporal consistency~\cite{tzaban2022stitch}. 
However, these methods typically rely on face alignment and cropping as preprocessing.
Although StyleGAN3~\cite{karras2021alias} was introduced to support unaligned face inputs, a subsequent study~\cite{alaluf2022third} has shown that it struggles to encode facial features effectively without preprocessing, often resulting in structural artifacts. 
To overcome these limitations, methods such as VToonify~\cite{yang2022vtoonify} and StyleGANEX~\cite{yang2023styleganex} have been proposed to directly process videos beyond pre-aligned cropping. 
Nevertheless, these methods remain limited to 2D representations and have yet to be extended to 3D applications.

\subsection{3D Head Avatar}
Since the introduction of NeRF~\cite{mildenhall2021nerf}, implicit representation-based methods~\cite{yenamandra2021i3dmm,zheng2022avatar,hong2022headnerf,chan2022efficient,xu2023avatarmav} for head reconstruction have achieved remarkable progress. 
3DGS~\cite{kerbl20233d} has obtained a significant breakthrough in 3D reconstruction, further advancing the development of downstream applications such as 3D head modeling.
Although several Gaussian-based methods~\cite{qian2024gaussianavatars,xu2024gaussian,chen2024monogaussianavatar,ma20243d,xiang2024flashavatar,abdal2024gaussian,kirschstein2024gghead} have demonstrated high-quality head reconstruction and impressive rendering performance, they typically focus on photo-realistic avatars, with relatively limited exploration of avatar stylization.
Stylized head avatars, characterized by geometric abstraction and artistic expression, differ significantly from the photo-realistic avatars synthesized by the aforementioned methods.

Pre-trained 3D GANs~\cite{wu2016learning} enable high-quality generation, making 3D head stylization possible. 
Although fine-tuning 3D generators for geometric and texture-based stylization has proven effective~\cite{or2022stylesdf,jin2022dr,abdal20233davatargan,lan2023self,wang2023rodin,zhang2024rodinhd}, performing independent fine-tuning for each new style remains costly. 
Toonify3D~\cite{jang2024toonify3d} addressed this limitation by predicting facial surface normals using the proposed StyleNormal, enabling direct face stylization without additional fine-tuning. 
Similarly, DeformToon3D~\cite{zhang2023deformtoon3d} introduced StyleField to predict conditional 3D deformations, aligning NeRF representations in real space with style space to achieve geometric stylization and obviate per-style fine‑tuning. 
However, Toonify3D suffers from limited data diversity, and DeformToon3D cannot support novel-view animations, which limits their application scenarios.

\section{Method}
\subsection{ToonifyGB Framework}
Given a monocular video input, ToonifyGB applies frame-by-frame stylization to generate the corresponding stylized frames. 
For inputs such as live streams or selfie videos, the face often occupies only a small portion of each frame, while the rest includes the hairstyle and upper body. 
Traditional methods~\cite{karras2019style,karras2020analyzing} typically require face alignment, cropping, and editing before synthesizing the results back into the original frame.
This process often introduces visual discontinuities at the seams, resulting in noticeable jitter in the output video. 
To address this issue, we adopt an improved StyleGAN model based on StyleGANEX~\cite{yang2023styleganex} in Stage~1, enabling stable stylized video generation at the original resolution, as shown in Figure~\ref{fig:pipeline}.

To prepare the training data for Stage~2, we follow the method in~\cite{zielonka2023instant,ma20243d}, using the facial tracker from~\cite{zielonka2022towards} to compute FLAME~\cite{li2017learning} meshes, including a neutral head model and a set of expression blendshapes.
This process also provides camera parameters $C$, joint and pose parameters $\Theta$, and expression coefficients $\{\psi_k\}$ for each frame.
In addition to enabling facial expressions control, the FLAME model based on Principal Component Analysis (PCA) provides joint and pose parameters for controlling head, eyeball, eyelid, and jaw movements.
As shown in Figure~\ref{fig:pipeline}, we apply Linear Blend Skinning (LBS) to transform the Gaussian model based on the extracted joint and pose parameters. 
The transformation is defined as:
\begin{equation}
B^{\psi*} = LBS(B^\psi, \Theta).
\label{eq:lbs}
\end{equation}
The transformed Gaussian model is then rendered in real-time as a 3D stylized head avatar using Gaussian Splatting.  
Finally, by integrating the camera parameters, we enable novel-view rendering and animation.

\begin{figure}[t]
\centering
\includegraphics[width=\linewidth]{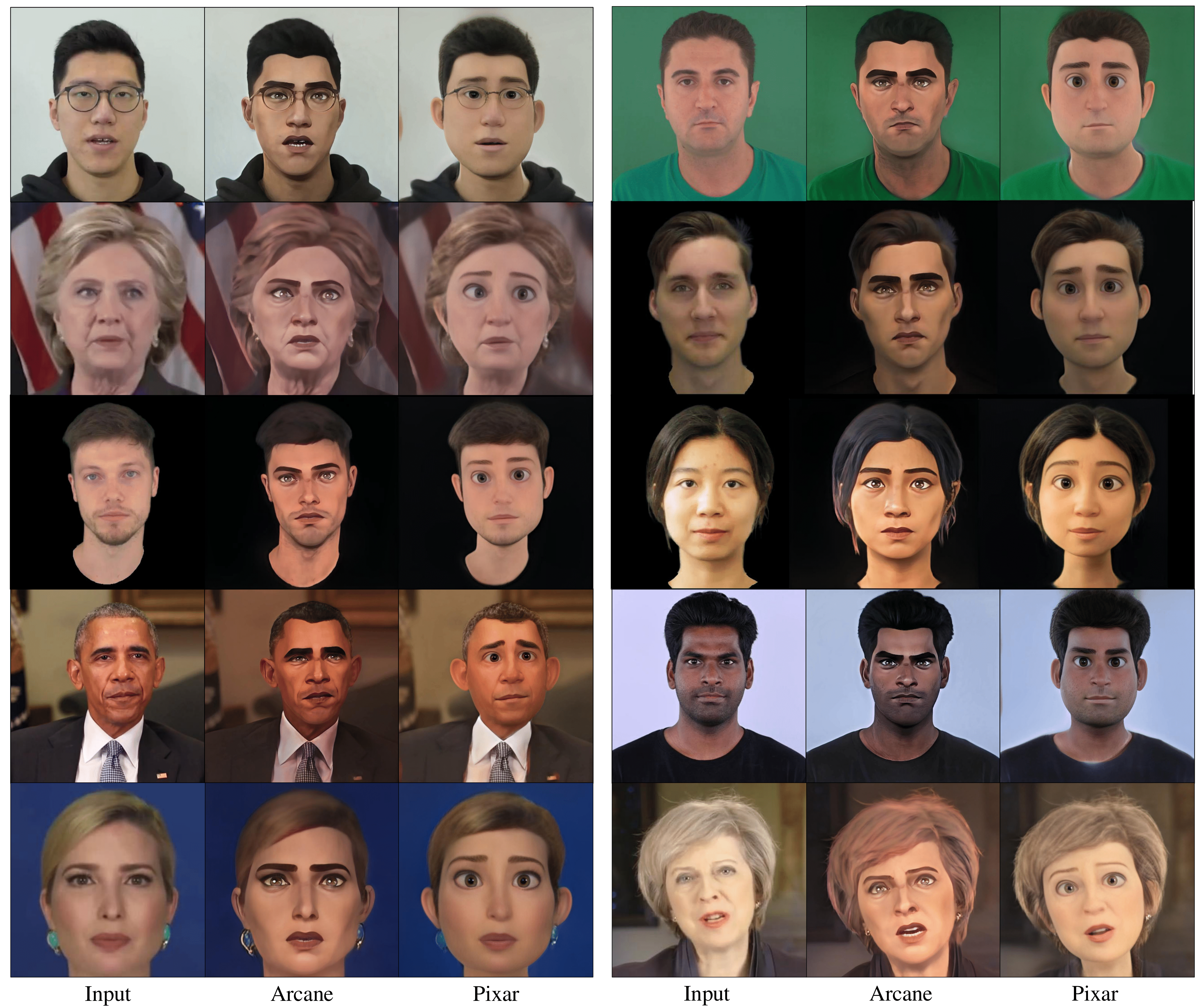}
\caption{\textbf{Visualization of stylized video generation results} in ``Arcane'' and ``Pixar'' styles on the INSTA~\cite{zielonka2023instant} and NeRFBlendShape~\cite{gao2022reconstructing} datasets, covering both male and female subjects.}
\label{fig:video}
\end{figure}

\subsection{Stylized Video Generation}
As shown in StyleGANEX~\cite{yang2023styleganex}, manipulating feature maps at different layers of StyleGAN leads to different spatial effects in the generated faces.
Specifically, while shifting or rotating the feature maps in deeper layers (i.e., Layer~7) produces consistent global transformations, similar operations in shallow layers (i.e., Layer~1) fail to preserve facial structure due to the low spatial resolution of the $4\times4$ feature map, causing blurring and loss of detail.
To address this limitation, we adopt StyleGANEX~\cite{yang2023styleganex}, an enhanced variant of StyleGAN2~\cite{karras2019style}, which increases the spatial resolution of shallow feature maps (Layers~1–7) to $32\times32$.
This improvement enables finer control over facial geometry and enhances the generation quality for unaligned faces.

Our specific architectural improvements of the generator are as follows. 
First, we replace the constant $4\times4$ input of the first layer with a variable feature map of resolution $1/32$ of the final output, enabling support for arbitrary input sizes.
Then, we replace the standard convolutions in the shallow layers with dilated convolutions to enlarge the receptive field. 
Finally, we remove all upsample operations before the eighth layer, ensuring that the seven shallow layers maintain the same $32\times32$ resolution.

These architectural improvements effectively address the limitations beyond pre-aligned cropping.
As shown in Figure~\ref{fig:video}, our method consistently generates high-quality stylized head videos across diverse styles, regardless of gender.

\subsection{Gaussian Blendshapes Synthesis}
We represent all Gaussian head avatars using 3D Gaussians.
As described in 3DGS~\cite{kerbl20233d}, each Gaussian has some basic properties including Gaussian center $\mu$, scale $s$, color $c$, opacity $\alpha$, and rotation $q$.
Based on 3DGB~\cite{ma20243d}, our Gaussian blendshape representation consists of a neutral model $B_0$ and a set of $n$ expression blendshapes ${B_1, B_2, \ldots, B_n}$.
Each Gaussian in the neutral model $B_0$ has a set of blend weights $w$ to control joint and pose.
In addition, each Gaussian in an expression blendshape $B_k$ corresponds one‑to‑one to a Gaussian in the neutral model $B_0$.
The difference between $B_k$ and $B_0$ is defined as the difference in their corresponding Gaussian properties: $\Delta B_k = B_k - B_0$.
The expression of Gaussian head avatar $B^\psi$ can be computed as follows:
\begin{equation}
B^\psi = B_0 + \sum_{k=1}^n \psi_k \,\Delta B_k
\end{equation}
where $\psi_k$ denotes the expression coefficients.
Here, $B^\psi$ represents the untransformed expression model, and the final posed Gaussian model, obtained via Linear Blend Skinning (LBS), is defined in Equation~\ref{eq:lbs}.

Since the FLAME meshes and blendshape models do not include interior mouth components such as teeth, we adopt the method of 3DGB~\cite{ma20243d} by defining a separate set of Gaussians for the mouth $B_m$. 
The properties of these mouth Gaussians are not affected by expression changes, they only move with the jaw joint in the FLAME model. 
The mouth Gaussians for the head avatar, $B_m^*$, are computed via linear blend skinning (LBS) as:
\begin{equation}
B_m^* = LBS(B_m, \Theta).
\end{equation} 
The transformed Gaussian model $(B^{\psi*}, B_m^*)$ is rendered into a complete 3D head avatar using real-time Gaussian Splatting, with the overall pipeline shown in Figure~\ref{fig:pipeline}.

\begin{figure*}[t]
\centering
\includegraphics[width=\linewidth]{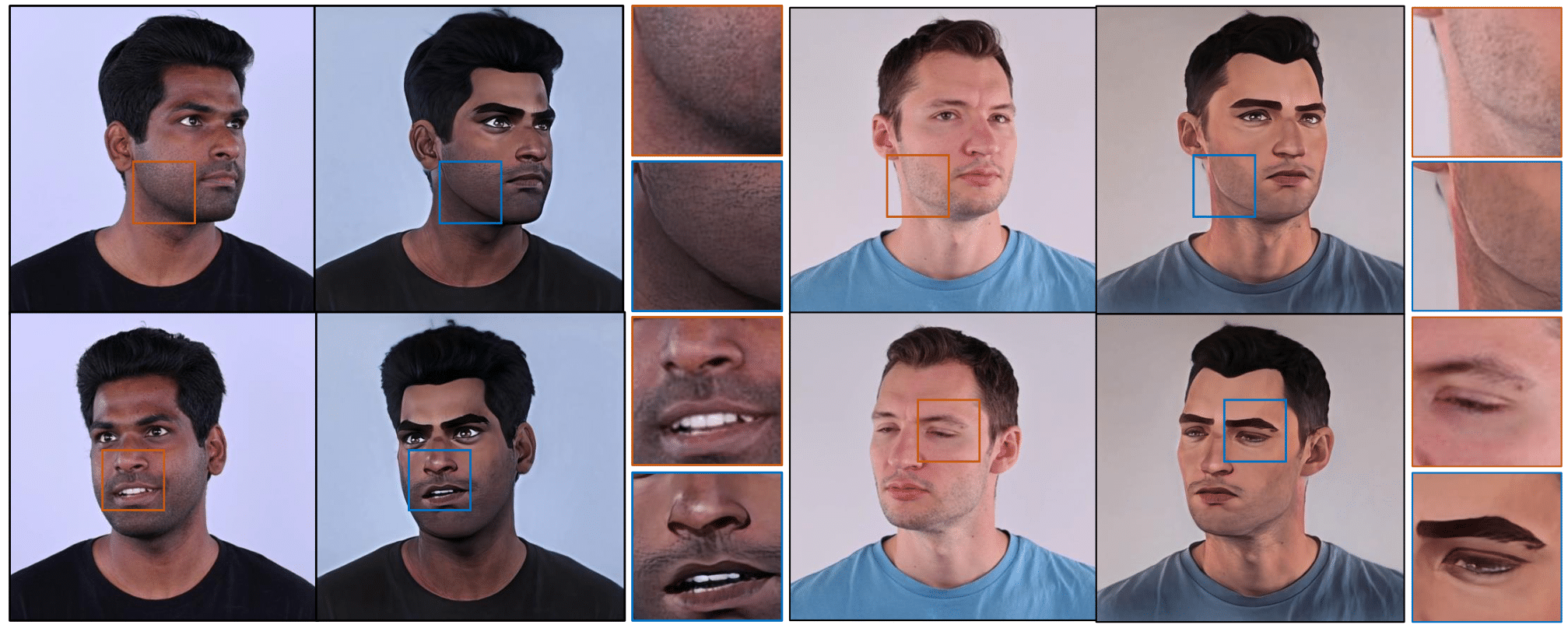}
\caption{\textbf{Visualization of stylized video generation results:} 
We present details of the real head from the input video, and the ``Arcane'' stylized head generated by our method. 
From left to right, the results for the video samples ``bala'' and ``wojtek\_1'' are shown.}
\label{fig:video_comparison}
\end{figure*}

\subsection{Loss Function}
We adopt the loss function from 3DGB~\cite{ma20243d}, and define the total loss as follows:
\begin{equation}
L = \lambda_1 L_{rgb} + \lambda_2 L_{\alpha} + \lambda_3 L_{reg},
\end{equation}
where the default weights of $\lambda_1$, $\lambda_2$ and $\lambda_3$ are set to 1, 10, 100, respectively.

The RGB loss $L_{rgb}$ encourages the rendered image to resemble the target video frame in both color and structure. 
It is computed as a weighted combination of an $L_1$ loss and a differentiable Structural Similarity (D-SSIM) loss:
\begin{equation}
L_{rgb} = \lambda_{rgb} L_1 + (1-\lambda_{rgb}) L_{D-SSIM},
\end{equation}
where the default weight $\lambda_{rgb}$ is set to $0.2$.

The opacity loss $L_{\alpha}$ penalizes opacity values outside the head mask. 
For each frame $i$, we compute the accumulated opacity image $I_{\alpha}^i$ and the corresponding head mask $M_h^i$, and average the error over $F$ frames:
\begin{equation}
L_{\alpha} = \frac{1}{F} \sum_{i=1}^{F} \frac{1}{|P|} \sum_{p \in P} \big( I_{\alpha}^i(p) - M_h^i(p) \big)^2.
\end{equation}

The regularization loss $L_{reg}$ constrains the mouth Gaussians to remain within a predefined cylindrical volume $V$. 
Let $\{\mathbf{x}_i\}_{i=1}^N$ denote the centers of Gaussians located in the mouth region. 
To penalize points outside the volume, we employ a signed distance function $SDF(\mathbf{x}_i, V)$, and define the loss as follows:
\begin{equation}
L_{reg} = \frac{1}{N} \sum_{i=1}^{N} \left( \max\left( SDF(\mathbf{x}_i, V), 0 \right) \right)^2,
\end{equation}
where $N$ is the number of mouth Gaussians.

\begin{table}[t]
\centering
\caption{\textbf{Video durations and inference times:} 
Duration (in seconds) of the input videos, and inference time (in seconds) of our method.}
\resizebox{\linewidth}{!}{
\begin{tabular}{c|cccccc}
\hline
Video samples & justin & malte\_1 & nf\_01 & bala & wojtek\_1 & person\_0004 \\ \hline
Duration & 98 & 130 & 130 & 159 & 137 & 60 \\
Inference & 221 & 260 & 213 & 342 & 275 & 108 \\ \hline
\end{tabular}}
\label{tab:inference}
\end{table}

\section{Experiments}
\subsection{Baselines}
Due to the current lack of methods for synthesizing 3D stylized head avatars using Gaussian blendshapes, we compare our method against the following state-of-the-art methods for photo-realistic 3D head avatar synthesis: 
INSTA~\cite{zielonka2023instant}, PointAvatar~\cite{zheng2023pointavatar}, FLARE~\cite{bharadwaj2023flare}, SplattingAvatar~\cite{shao2024splattingavatar}, FlashAvatar~\cite{xiang2024flashavatar}, and 3DGB~\cite{ma20243d}.
Notably, 3DGB shares a similar architecture with ours but focuses on photo-realistic avatar synthesis and does not support the synthesis of diverse stylized avatars.

\begin{table}[t]
\centering
\caption{\textbf{Quantitative comparison of video stabilization:} 
We compare the original input (OI), the aligned input (AI), our ``Arcane'' (OA), and the aligned ``Arcane'' (AA) videos.}
\resizebox{\linewidth}{!}{
\begin{tabular}{c|c|cccccc}
\hline
\multicolumn{2}{c|}{Video Samples} & justin & malte\_1 & nf\_01 & bala & wojtek\_1 & person\_0004 \\ \hline
\multirow{4}{*}{ITF$\uparrow$} & OI & 37.78 & 38.47 & 31.82 & 37.73 & 39.02 & 37.17 \\
 & AI & 32.45 & 28.84 & 26.49 & 27.97 & 29.09 & 34.80 \\ \cline{2-8}
 & OA & 35.80 & 34.51 & 29.36 & 36.01 & 36.77 & 33.82 \\
 & AA & 31.35 & 26.04 & 25.31 & 26.43 & 28.31 & 30.84 \\ \hline
\multirow{4}{*}{ISI$\uparrow$} & OI & 0.9685 & 0.9709 & 0.9361 & 0.9614 & 0.9651 & 0.9277 \\
 & AI & 0.9066 & 0.9276 & 0.8918 & 0.8995 & 0.9126 & 0.9270 \\ \cline{2-8}
 & OA & 0.9700 & 0.9643 & 0.9382 & 0.9685 & 0.9670 & 0.9532 \\
 & AA & 0.9034 & 0.8965 & 0.8887 & 0.8963 & 0.9030 & 0.9036 \\ \hline
\end{tabular}}
\label{tab:jitter}
\end{table}

\subsection{Dataset}
We evaluate both our method and state-of-the-art photo-realistic avatar synthesis methods using six videos from the INSTA~\cite{zielonka2023instant} dataset. 
Each video is cropped and resized to $512 \times 512$ resolution, with sequence lengths ranging from 1,000 to 4,000 frames.
Following the method of 3DGB~\cite{ma20243d}, we retain the final 350 frames of each video for testing.
Both 3DGB~\cite{ma20243d} and our method apply the same preprocessing pipeline~\cite{zielonka2022towards,zielonka2023instant}, including background removal and FLAME parameter extraction.

\begin{table*}[t]
\centering
\caption{\textbf{Quantitative comparison of 3D head avatars:} 
We evaluate our method and state-of-the-art methods on the INSTA~\cite{zielonka2023instant} dataset. 
In each metric group, the best value is highlighted in \textbf{bold}, and the second‑best is \underline{underlined}.
}
\resizebox{\linewidth}{!}{
\begin{tabular}{c|cc|cc|cc|cc|cc|cc}
\hline
\multirow{2}{*}{Method} & \multicolumn{2}{c|}{justin} & \multicolumn{2}{c|}{malte\_1} & \multicolumn{2}{c|}{nf\_01} & \multicolumn{2}{c|}{bala} & \multicolumn{2}{c|}{wojtek\_1} & \multicolumn{2}{c}{person\_0004} \\ \cline{2-13}
 & PSNR$\uparrow$ & SSIM$\uparrow$ & PSNR$\uparrow$ & SSIM$\uparrow$ & PSNR$\uparrow$ & SSIM$\uparrow$ & PSNR$\uparrow$ & SSIM$\uparrow$ & PSNR$\uparrow$ & SSIM$\uparrow$ & PSNR$\uparrow$ & SSIM$\uparrow$ \\ \hline
INSTA~\cite{zielonka2023instant} & 31.66 & 0.9591 & 27.44 & 0.9159 & 26.45 & 0.8937 & 29.53 & 0.8896 & \underline{31.36} & 0.9452 & 25.44 & 0.8478 \\
PointAvatar~\cite{zheng2023pointavatar} & 30.40 & 0.9373 & 24.98 & 0.8853 & 25.25 & 0.8919 & 27.88 & 0.8658 & 28.82 & 0.9192 & 23.29 & 0.8576 \\
FLARE~\cite{bharadwaj2023flare} & 29.10 & 0.9363 & 25.93 & 0.8973 & 25.97 & 0.9027 & 27.20 & 0.8761 & 27.84 & 0.9216 & 25.53 & 0.9015 \\
SplattingAvatar~\cite{shao2024splattingavatar} & 30.93 & 0.9482 & 27.66 & 0.9243 & 27.08 & 0.9202 & 32.14 & 0.9272 & 29.54 & 0.9400 & \underline{26.49} & \underline{0.9075} \\
FlashAvatar~\cite{xiang2024flashavatar} & 32.16 & 0.9611 & 27.45 & 0.9326 & 28.02 & 0.9326 & 30.27 & 0.8494 & 32.02 & 0.9509 & 25.49 & 0.8996 \\
3DGB~\cite{ma20243d} & 32.63 & \underline{0.9643} & \underline{28.65} & \textbf{0.9432} & 28.06 & \underline{0.9340} & \underline{33.29} & \underline{0.9457} & \textbf{32.57} & \textbf{0.9623} & 23.66 & 0.8449 \\
Ours (Arcane) & \underline{33.12} & 0.9628 & \textbf{29.55} & 0.9360 & \underline{28.33} & 0.9288 & \textbf{33.39} & \textbf{0.9488} & 30.56 & 0.9436 & \textbf{28.76} & \textbf{0.9110} \\
Ours (Pixar) & \textbf{33.42} & \textbf{0.9662} & 27.01 & \underline{0.9375} & \textbf{28.34} & \textbf{0.9341} & 30.84 & 0.9337 & 31.14 & \underline{0.9583} & 23.16 & 0.8338 \\ \hline
\end{tabular}}
\label{tab:comparison}
\end{table*}

\begin{table}[t]
\centering
\caption{\textbf{Performance comparison:}
We record the training time (in minutes) and the rendering speed (in fps) of 3DGB and our method in both ``Arcane'' (A) and ``Pixar'' (P) styles.}
\resizebox{\linewidth}{!}{
\begin{tabular}{c|c|cccccc}
\hline
\multicolumn{2}{c|}{Video Samples} & justin & malte\_1 & nf\_01 & bala & wojtek\_1 & person\_0004 \\ \hline
\multirow{3}{*}{Train$\downarrow$} & 3DGB & 41 & 44 & 44 & \textbf{44} & 49 & 45 \\
 & Ours (A) & \textbf{40} & 45 & 44 & 45 & 50 & \textbf{44} \\
 & Ours (P) & 43 & \textbf{40} & \textbf{43} & \textbf{44} & \textbf{45} & \textbf{44} \\ \hline
\multirow{3}{*}{Render$\uparrow$} & 3DGB & \textbf{143} & \textbf{142} & 130 & 134 & 138 & \textbf{134} \\
 & Ours (A) & 140 & \textbf{142} & \textbf{131} & \textbf{135} & \textbf{140} & 128 \\
 & Ours (P) & 141 & 133 & 128 & 132 & 134 & 127 \\ \hline
\end{tabular}}
\label{tab:time}
\end{table}

\subsection{Evaluation Metrics}
We employ two metrics to evaluate video stabilization: 
Inter-frame Transformation Fidelity (ITF)~\cite{morimoto1998evaluation,marcenaro2001image,xu2012fast} and Inter-frame Similarity Index (ISI)~\cite{guilluy2018performance,james2023globalflownet}.
ITF measures the inter-frame Peak Signal-to-Noise Ratio (PSNR) in dB based on the mean squared error.
The intuitive idea of ITF is that a more stable video (i.e., less jittery) will have greater similarity between adjacent frames compared to an unstable version of the same video.
ISI computes the average Structural Similarity (SSIM) between adjacent frames across the video.
Higher ISI values indicate greater perceptual similarity between frames, leading to improved visual comfort for viewers.

For 3D head avatar synthesis, we evaluate the performance of our method and state-of-the-art methods for photo-realistic avatar synthesis using standard evaluation metrics~\cite{zhang2018unreasonable,ma20243d}, including Peak Signal-to-Noise Ratio (PSNR) and Structural Similarity Index (SSIM).
In addition, we record the training time (in minutes) and the rendering speed (in frames per second, fps) for each method.
In the ablation study, we additionally adopt the Learned Perceptual Image Patch Similarity (LPIPS) metric to better capture perceptual differences between the synthesized avatars and the ground truth.

\subsection{Implementation Details}
To ensure a fair performance comparison, the training and testing of all methods are performed on a single RTX 4090 GPU. 
Our methods are implemented in Python using the PyTorch framework.

For 2D stylized video generation, we use the pre-trained models provided by StyleGANEX~\cite{yang2023styleganex}.
For training the 3D stylized head avatars, we employ the Adam optimizer~\cite{kingma2014adam}, setting the initial learning rates of the Gaussian properties $\{\mathbf{x}_{k},\alpha_{k}, \mathbf{s}_{k}, \mathbf{q}_{k}, S_{H_{k}}\}$ to $3.2\times10^{-7}$, $5\times10^{-5}$, $5\times10^{-4}$, $1\times10^{-4}$, and $1.25\times10^{-3}$, respectively. 
Following 3DGB~\cite{ma20243d}, the initial number of sampled Gaussians is 50k for the neutral head model and 14k for the mouth interior.

\subsection{Quantitative Comparison}
\subsubsection{Video Stabilization}
We adopt an improved StyleGAN model to generate six videos in the ``Arcane'' style.  
The durations of the videos and their corresponding inference times are summarized in Table~\ref{tab:inference}.  
All input videos have a resolution of $512 \times 512$ pixels, and inference is performed on a single NVIDIA RTX 4090 GPU.  
For video durations ranging from 60 to 160 seconds, the generation times span approximately 100 to 350 seconds.

To evaluate the impact of preprocessing, we apply a standard face alignment technique based on a facial keypoint predictor~\cite{kazemi2014one} to the input videos.  
We compare the original input videos (Original Input, OI) with their aligned counterparts (Aligned Input, AI).  
Likewise, we compare the ``Arcane'' style outputs generated from unaligned inputs (Ours Arcane, OA) with those generated from aligned inputs (Aligned Arcane, AA).  

As shown in Table~\ref{tab:jitter}, both the Inter-frame Transformation Fidelity (ITF) and Inter-frame Similarity Index (ISI) scores for AI are consistently lower than those for OI.  
Similarly, AA exhibits lower ITF and ISI scores compared to OA.  
These results suggest that applying face alignment and cropping prior to frame-by-frame generation (i.e., AI and AA) tends to introduce greater temporal instability, resulting in more jittery outputs.

\begin{figure}[t]
\centering
\includegraphics[width=\linewidth]{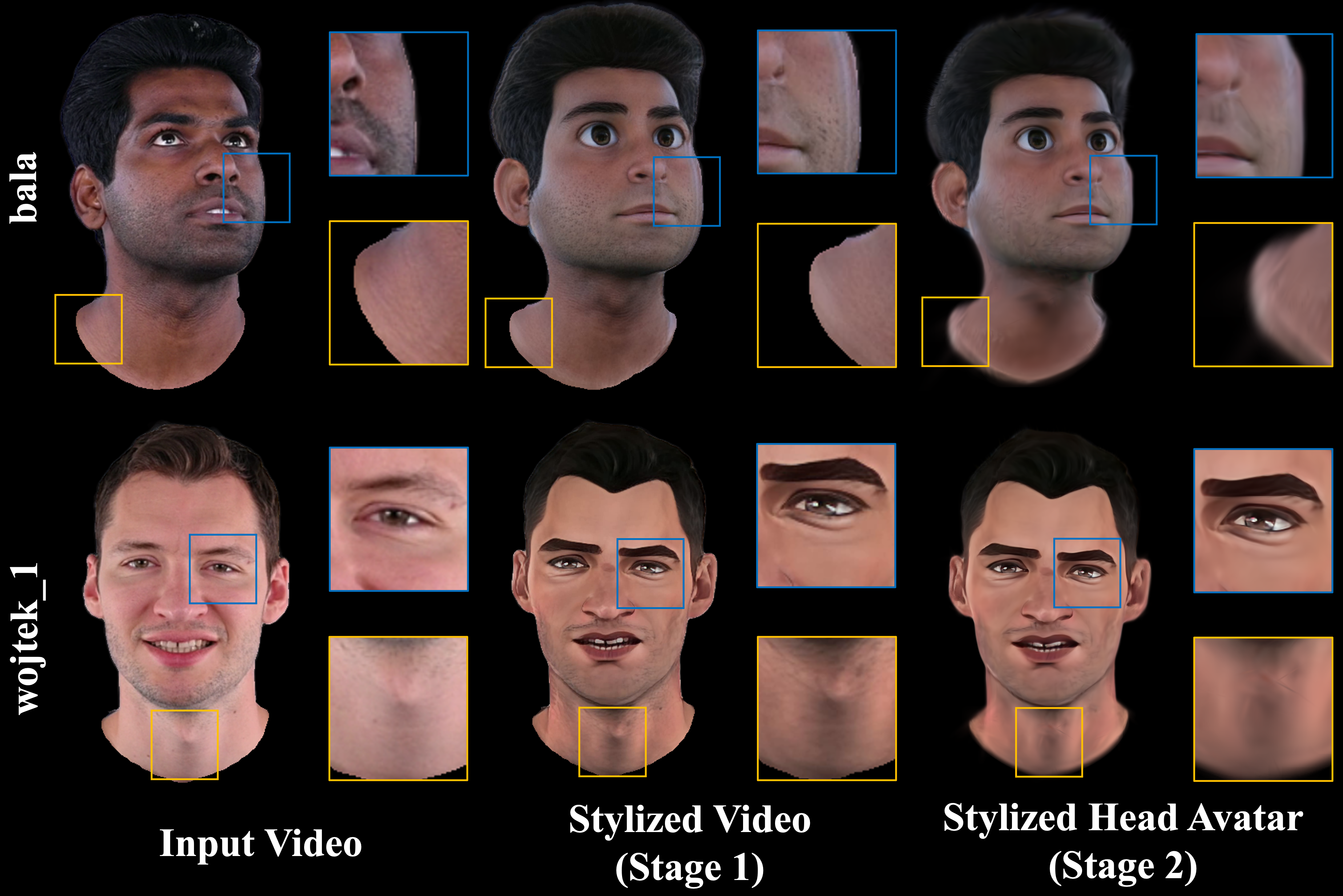}
\caption{\textbf{Qualitative comparison of each stage:} 
We present the input video head frames, the corresponding stylized videos, and 3D head avatars synthesized by our method.}
\label{fig:qualitative2}
\end{figure}

\begin{figure}[t]
\centering
\includegraphics[width=\linewidth]{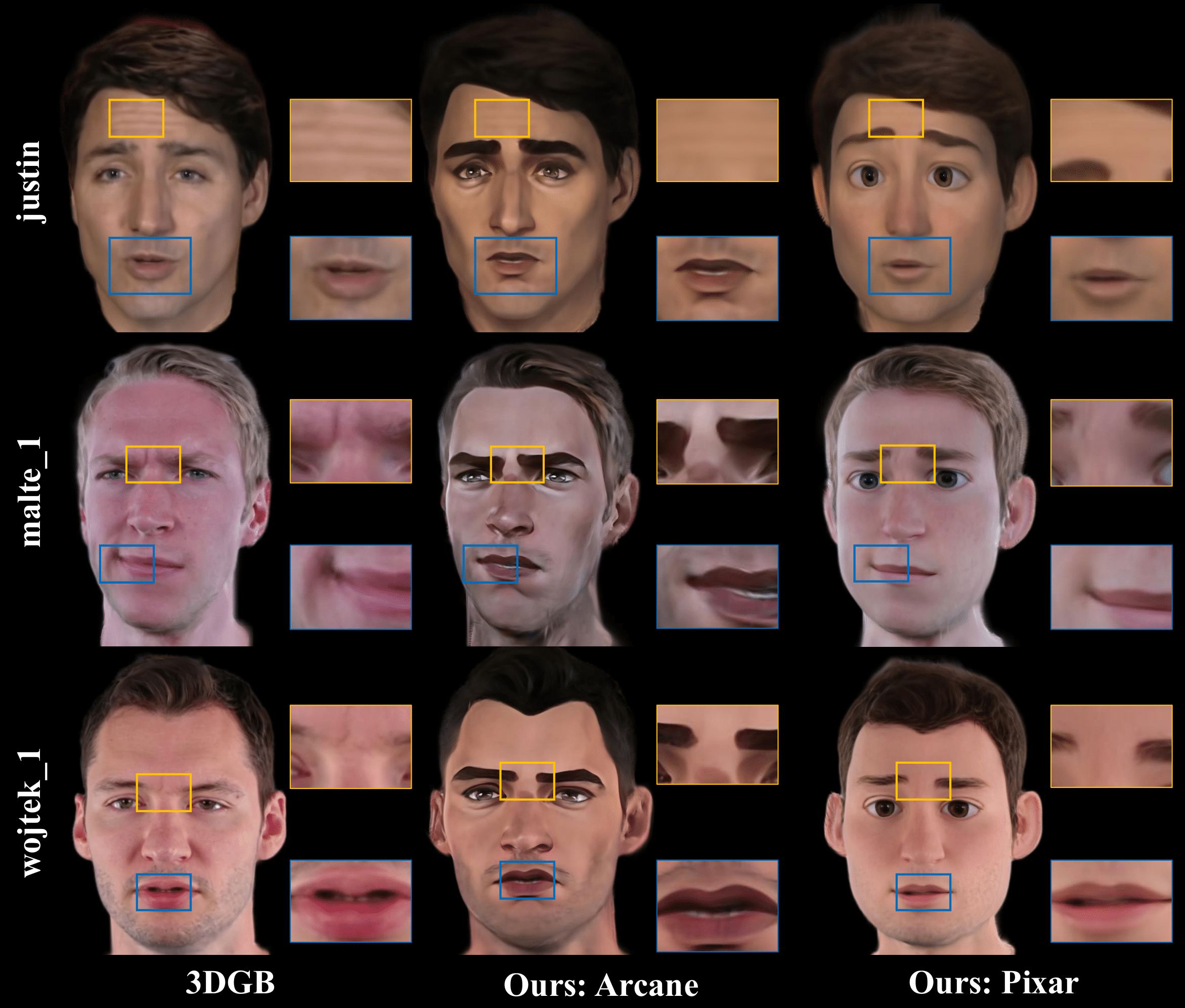}
\caption{\textbf{Qualitative comparison of baseline and ours:} 
We present 3D head avatars using Gaussian blendshapes synthesized by 3DGB~\cite{ma20243d} and our method.}
\label{fig:qualitative}
\end{figure}

\begin{table*}[t]
\centering
\caption{\textbf{User preference study:} 
We conduct user preference studies on the ``Arcane'' and ``Pixar'' styles, where users rate their preferences on a scale from 1 to 5, with higher scores indicating greater satisfaction, across three evaluation criteria: Style Consistency, Identity Preservation, and Overall Quality.
The highest percentage is highlighted in \textbf{bold}.}
\resizebox{\linewidth}{!}{
\begin{tabular}{l|ccc|ccc}
\hline
Style & \multicolumn{3}{c|}{Arcane} & \multicolumn{3}{c}{Pixar} \\ \hline
Evaluation Criteria & Style Consistency & Identity Preservation & Overall Quality & Style Consistency & Identity Preservation & Overall Quality \\ \hline
1: Very Dissatisfied & 0.0\% & 2.8\% & 0.6\% & 2.8\% & 5.6\% & 2.2\% \\
2: Dissatisfied & 5.0\% & 10.0\% & 6.1\% & 5.0\% & 6.7\% & 5.0\% \\
3: Neutral & 11.1\% & 15.6\% & 15.0\% & 13.3\% & 24.4\% & 20.0\% \\
4: Satisfied & 32.2\% & \textbf{39.4\%} & 36.7\% & 29.4\% & \textbf{36.7\%} & \textbf{37.8\%} \\
5: Very Satisfied & \textbf{51.7\%} & 32.2\% & \textbf{41.7\%} & \textbf{49.4\%} & 26.7\% & 35.0\% \\ \hline
\end{tabular}}
\label{tab:user}
\end{table*}

\subsubsection{3D Head Avatar}
We evaluate our method and state‑of‑the‑art methods using standard metrics for animatable head reconstruction. 
The quantitative results are presented in Table~\ref{tab:comparison}, and the training and rendering times for both the baseline methods and ours are reported in Table~\ref{tab:time}.
With the additional integration of stylization, our method achieves performance comparable to the state‑of‑the‑art on the PSNR and SSIM metrics in most cases, and even outperforms them on certain data. 
Specifically, our method outperforms all other methods on synthesizing the ``Arcane'' style for the ``bala'' and ``person\_0004'' data, as well as the ``Pixar'' style for the ``justin'' and ``nf\_01'' data.

In addition, although our method integrates stylization into 3D head avatars, its training and rendering times remain comparable to those of the method of 3DGB~\cite{ma20243d}. 
In certain cases, our method is even more efficient in both training and rendering. 
Combined with the additional time required for video generation (as shown in Table~\ref{tab:inference}), the overall time cost of our method remains acceptable.

\subsection{Qualitative Comparison}
We present the original video head frames, the corresponding stylized video frames generated by our method, and the 3D stylized head avatars synthesized using Gaussian blendshapes. 
The qualitative comparison is shown in Figure~\ref{fig:qualitative2}. 
The examples are selected from the ``bala'' dataset in the ``Pixar'' style and the ``wojtek\_1'' dataset in the ``Arcane'' style.

In the stylized video, the ``bala'' data exhibits artifacts along the side edge of the head. 
We attribute this to the latent space distribution learned by StyleGAN, which tends to produce striped artifacts when the viewing angle falls outside the distribution covered by the training data. 
Notably, these artifacts are not present in the corresponding 3D stylized head avatars rendered by our method. 
Furthermore, the 3D stylized head avatars successfully preserve fine details from the stylized videos, such as the mole near the eye in the ``wojtek\_1'' dataset.
However, since the 3D avatar synthesis mainly focuses on the facial region, the neck area is typically blurred, as observed in both cases. 
This blurring leads to the lower quantitative performance, since the neck region is included in the evaluation.

The qualitative comparison with 3DGB~\cite{ma20243d} is presented in Figures~\ref{fig:qualitative}. 
Our method effectively captures and preserves high-frequency details in the stylized videos. 
Compared to the state-of-the-art method, ToonifyGB can synthesize 3D stylized head avatars with comparable quality and detail.

\subsection{Visualization}
To better demonstrate the visual quality of our generated videos, we present several examples in Figure~\ref{fig:video}, and select two representative videos for detailed comparison in Figure~\ref{fig:video_comparison}. 
Specifically, we show real head frames from the ``bala'' and ``wojtek\_1'' videos, as well as the corresponding heads of generated videos in the ``Arcane'' style.

The results demonstrate that key facial features, such as the beard, mouth shape, and even small details like the black mole above the eye in the lower right image, are well preserved after the stylization process. 
These details highlight the excellent performance of our method in terms of detail preservation and identity consistency.

\begin{table}[t]
\centering
\caption{\textbf{Ablation study on face alignment and cropping:} 
We compare 3D head avatars synthesized from different input videos: 
one generated by our method, and the other using face alignment and cropping as prprocessing.}
\resizebox{\linewidth}{!}{
\begin{tabular}{l|ccc}
\hline
Method & PSNR$\uparrow$ & SSIM$\uparrow$ & LPIPS$\downarrow$ \\ \hline
Face Align \& Crop & 32.23 & 0.9387 & 0.1587 \\
Ours & 33.27 & 0.9645 & 0.0796 \\ \hline
\end{tabular}}
\label{tab:ablation}
\end{table}

\begin{figure}[t]
\centering
\includegraphics[width=\linewidth]{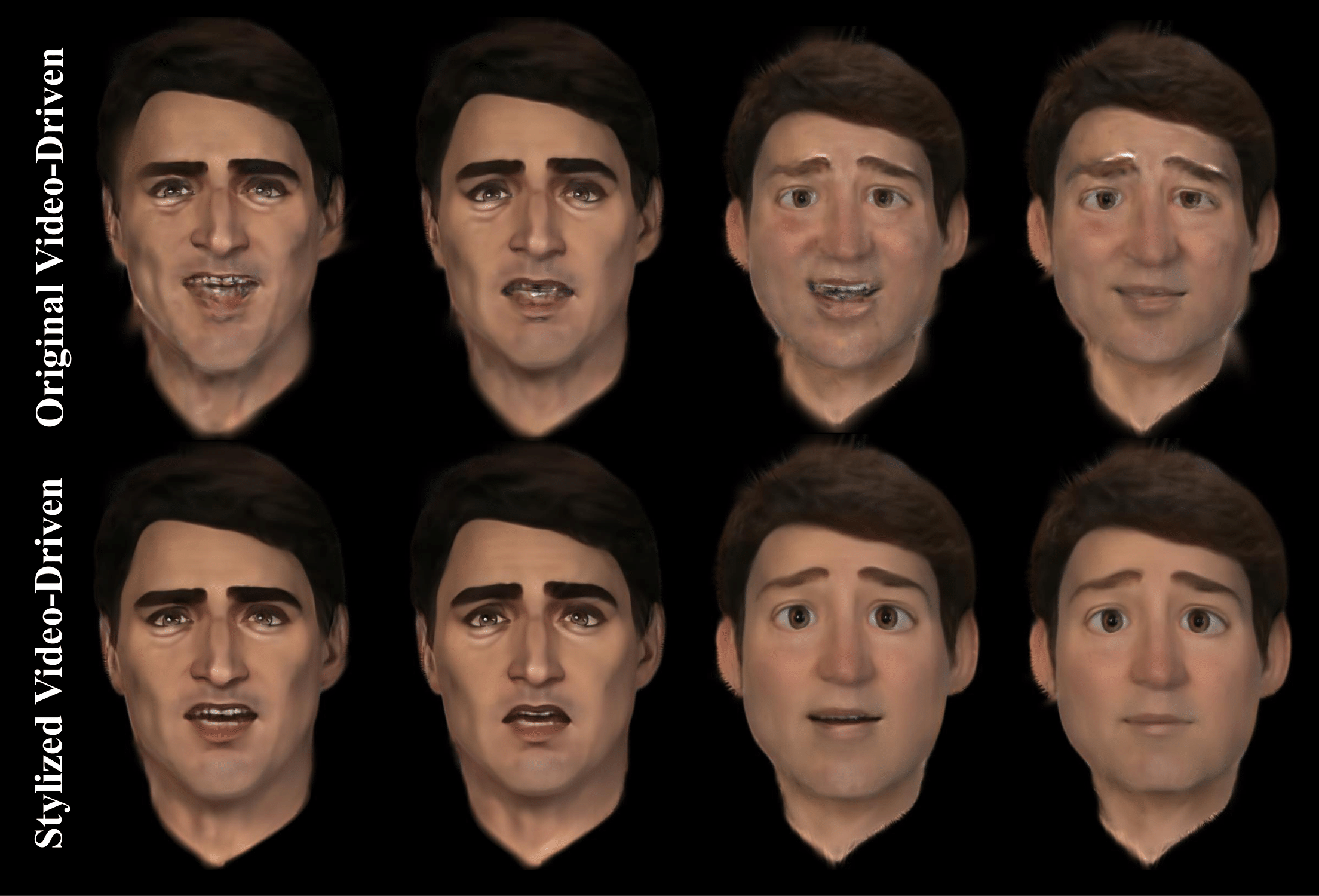}
\caption{\textbf{Ablation study on the effect of different driving videos:}
We present 3D stylized head avatar animation driven by the original input videos and our generated videos.}
\label{fig:ablation}
\end{figure}

\subsection{User Study}
To more effectively evaluate the 3D stylized head avatars synthesized by our method, we conduct a user preference study. 
We collect 180 votes for both the ``Arcane'' and ``Pixar'' styles, respectively, with users rating their preferences using Likert scales~\cite{likert1932technique} across three evaluation criteria: 
Style Consistency, Identity Preservation, and Overall Quality. 
Each criterion is assessed using a five-point scale: 1 for Very Dissatisfied, 2 for Dissatisfied, 3 for Neutral, 4 for Satisfied, and 5 for Very Satisfied. 

Specifically, Style Consistency evaluates how well the stylized output aligns with the defining characteristics of the style; 
Identity Preservation measures whether the avatar retains the unique features and identity of the original character after stylization; and Overall Quality provides a comprehensive assessment of the visual appeal and overall quality of the synthesized avatar.

The results are presented in Table~\ref{tab:user}, indicating that most users show great satisfaction with the ``Arcane'' style, particularly in terms of Style Consistency (51.7\% of users are very satisfied) and Overall Quality (41.7\% of users are very satisfied).
Although the rating for Identity Preservation is slightly lower, the average score remains favorable. 

In addition, the ``Pixar'' style is also favored by users, particularly in Style Consistency (49.4\% of users are very satisfied). 
For Identity Preservation and Overall Quality, the majority of users (over 60\%) indicate satisfaction with our 3D stylized head avatars.

\begin{figure}[t]
\centering
\includegraphics[width=\linewidth]{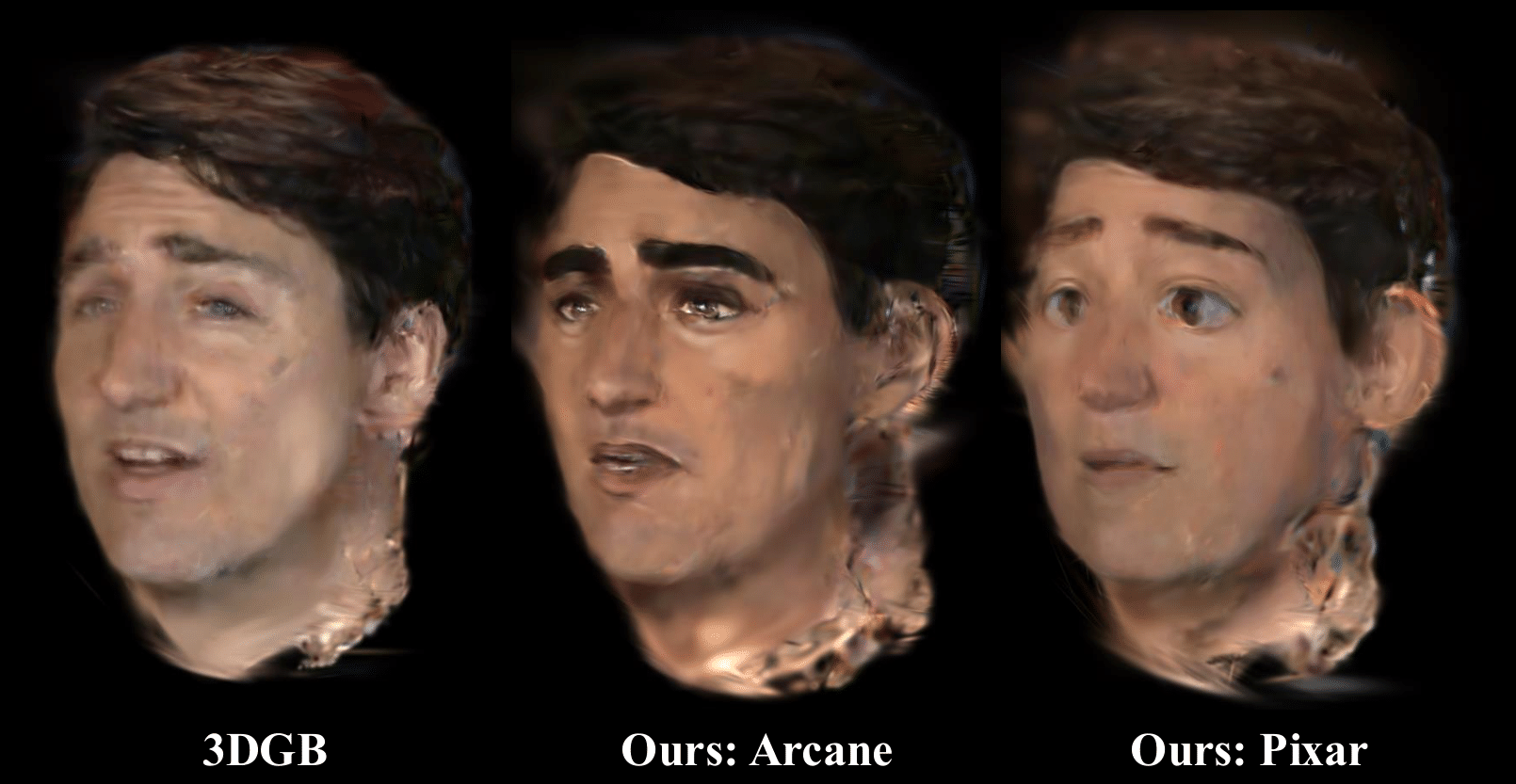}
\caption{\textbf{Limitation:} 
We present side-view renderings synthesized by 3DGB~\cite{ma20243d} and our method.}
\label{fig:limitations}
\end{figure}

\subsection{Ablation Study}
\subsubsection{Face Alignment and Cropping}
We compare 3D stylized head avatars (using the ``justin'' data) synthesized from videos processed by our method against those generated from videos preprocessed with face alignment and cropping. 
The resulting avatars are evaluated using PSNR, SSIM, and Learned Perceptual Image Patch Similarity (LPIPS). 
As shown in Table~\ref{tab:ablation}, our method outperforms the traditional method with face alignment and cropping across all evaluation metrics. 
This demonstrates that our method effectively eliminates jitter during video generation, enabling higher-quality synthesis of 3D stylized head animations.

\subsubsection{Source Videos for Driving Animation}
Compared with the architecture of 3DGB~\cite{ma20243d} that synthesizes 3D photo-realistic head avatars, our framework includes an additional Stage~1 to generate the stylized video. 
To demonstrate the importance of the generated stylized video in driving the animation, we compare the results of using the original input video (real face) versus our generated stylized video as the driving source, as shown in Figure~\ref{fig:ablation}.

It can be observed that using the original input video (real face) as the driving source often leads to unsatisfactory results, especially around the mouth region. 
This error occurs due to significant differences in expression blendshapes between the real and stylized domains. 
These results highlight the importance of the stylized videos generated by Stage~1 of our framework. 
Therefore, we recommend using the generated stylized videos, rather than the original input videos, as the driving source for 3D stylized head avatar animation. 

\section{Limitation}
Our method struggles to render side views of 3D stylized head avatars when the training data (i.e., input video) lacks side-view representations of the real head. 
As shown in Figure~\ref{fig:limitations}, we present side-view renderings synthesized by both 3DGB~\cite{ma20243d} and our method, and this limitation is also observed in the state-of-the-art methods. 
In fact, existing NeRF-based and Gaussian-based methods have yet to effectively address this issue. 
Rendering novel views from single-view training data remains an open problem for future research. 
Two directions to address this limitation include employing 2D GANs to synthesize videos with side views as additional training data, and enhancing the generalization ability of our model.

\section{Conclusion}
We propose a novel two-stage framework, named ToonifyGB, which utilizes Gaussian blendshapes to synthesize head animations in diverse styles from  monocular videos. 
In Stage~1, the proposed method adopts an improved StyleGAN-based model to generate stylized videos without requiring face alignment or cropping as preprocessing. 
This results in more temporally stable outputs, providing a reliable foundation for high-quality 3D head avatar animation synthesis.
Stage~2 focuses on constructing 3D stylized head avatars using Gaussian blendshapes, enabling fine-grained expression modeling and satisfactory animation. 
Our method supports real-time generation of stylized avatar animations in popular styles such as ``Arcane'' and ``Pixar''.

For future work, we plan to integrate motion capture technologies to enable real-time expression control of 3D stylized avatars. 
Specifically, we aim to explore more efficient approaches for obtaining real-time expression parameters, bypassing the complexity of traditional PCA inversion. 
This direction is expected to further broaden the applicability of ToonifyGB in virtual character interaction and personalized avatar generation.

\section*{Acknowledgments}
This work was supported in part by the National Science and Technology Council (NSTC), Taiwan, under Grants NSTC 114-2221-E-002-001- and NSTC 114-2420-H-002-010-.

{
\small
\bibliographystyle{ieeenat_fullname}
\bibliography{main}
}

\begin{figure*}[t]
\centering
\includegraphics[width=\linewidth]{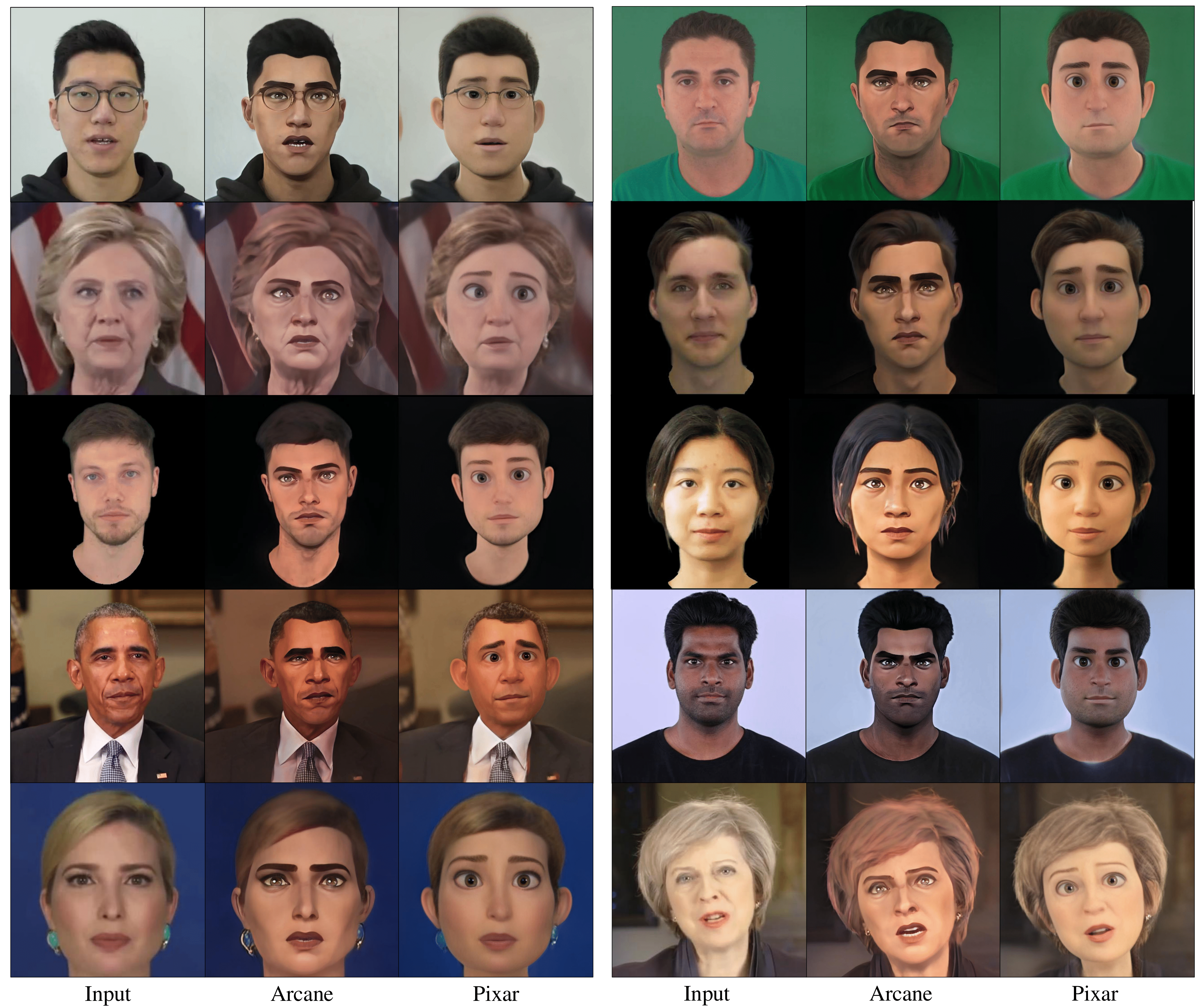}
\caption{Visualization of stylized video generation results on the videos from the INSTA~\cite{zielonka2023instant} and NeRFBlendShape~\cite{gao2022reconstructing} datasets.}
\end{figure*}

\begin{figure*}[t]
\centering
\includegraphics[width=\linewidth]{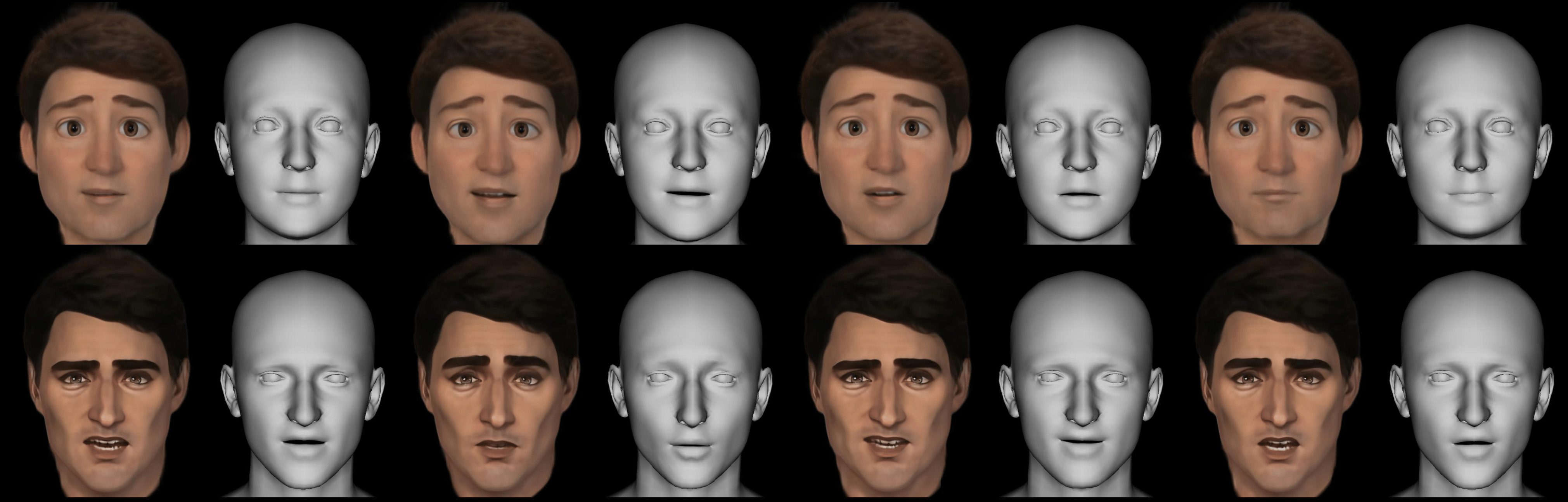}
\caption{Visualization of the synthesized 3D stylized head avatars. 
Each avatar closely resembles its corresponding FLAME mesh while capturing the stylized appearance.}
\end{figure*}

\begin{figure*}[t]
\centering
\includegraphics[width=\linewidth]{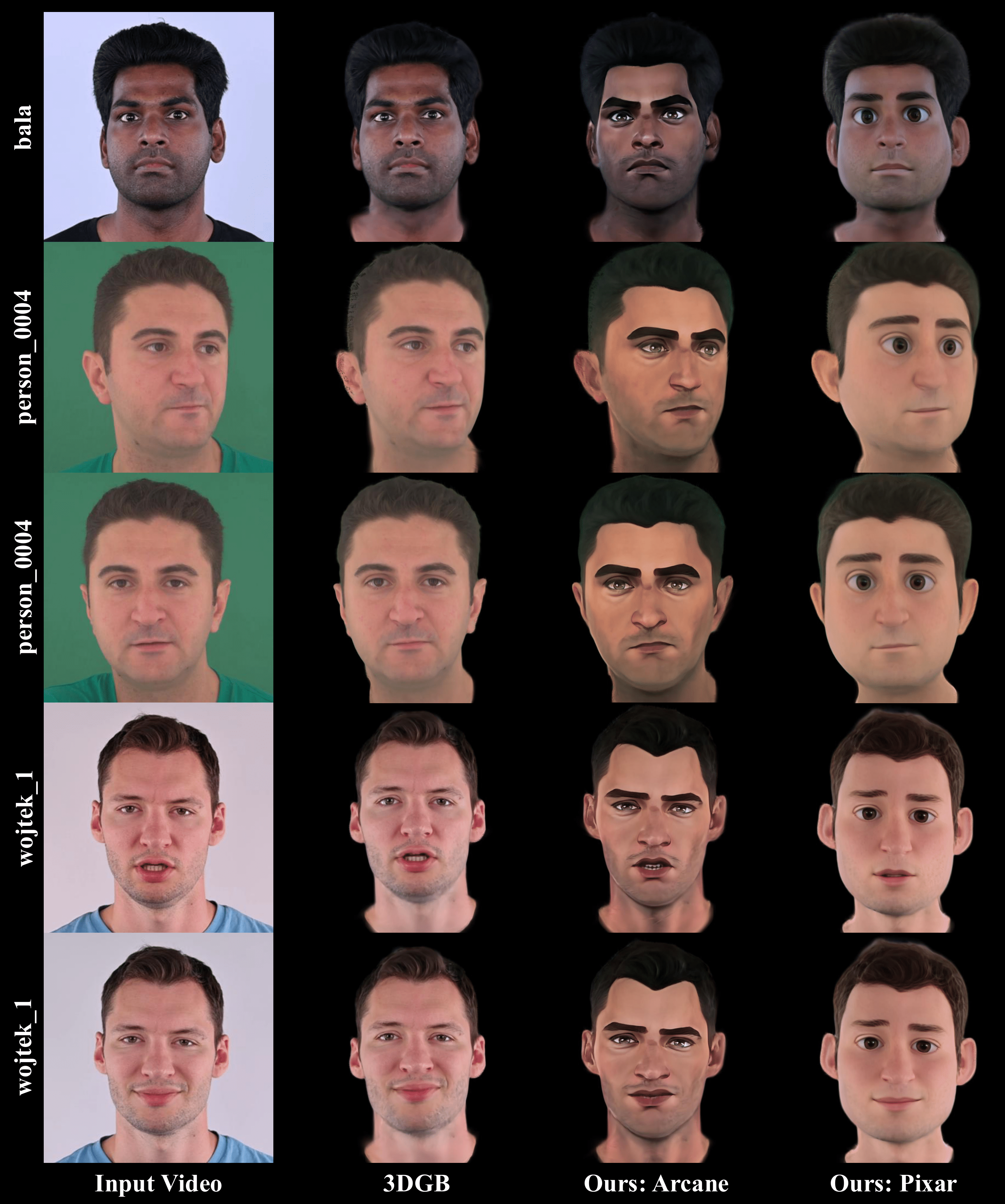}
\caption{\textbf{More examples for qualitative comparison:} 
We present input video head frames, and 3D head avatars using Gaussian blendshapes synthesized by 3DGB~\cite{ma20243d} and our method.}
\end{figure*}

\end{document}